\documentclass{svproc}
\usepackage{dirtytalk} 
\usepackage{type1cm}        
\usepackage{graphicx}        

\usepackage[bottom]{footmisc}
\usepackage{xcolor}
\usepackage{newtxmath}       
\usepackage{hyperref}

\usepackage[style=numeric]{biblatex}

\usepackage{url}

\begin{document}
	\mainmatter              
	\title{Semi-Supervised Learning guided by the Generalized Bayes Rule under Soft Revision}
	\titlerunning{SSL with Generalized Bayes}  
	%
	\author{Stefan Dietrich, Julian Rodemann and Christoph Jansen}
	\authorrunning{Stefan Dietrich et al.} 
	%
	%
	\institute{Department of Statistics, Ludwig-Maximilians-Universität (LMU), Munich, Germany}
	
	\maketitle              
	
	\begin{abstract}
We provide a theoretical and computational investigation of the Gamma-Maximin method with soft revision, which was recently proposed as a robust criterion for pseudo-label selection (PLS) in semi-supervised learning. Opposed to traditional methods for PLS we use credal sets of priors (\say{generalized Bayes}) to represent the epistemic modeling uncertainty. These latter are then updated by the Gamma-Maximin method with soft revision. We eventually select pseudo-labeled data that are most likely in light of the least favorable distribution from the so updated credal set. We formalize the task of finding optimal pseudo-labeled data w.r.t. the Gamma-Maximin method with soft revision as an optimization problem. A concrete implementation for the class of logistic models then allows us to compare the predictive power of the method with competing approaches. It is observed that the Gamma-Maximin method with soft revision can achieve very promising results, especially when the proportion of labeled data is low.\footnote{This work is partly based on Stefan Dietrich's Bachelor thesis \say{$\alpha$-cuts as robust method of pseudo label selection within the context of semi-supervised learning}. }
	\end{abstract}
\section{Introduction}
In this paper, we investigate Gamma-Maximin with soft revision for robust selection of pseudo-labeled data in semi-supervised learning (SSL) problems, as recently proposed in~\cite{RodemannIAL}. We use this introduction to collect those concepts from SSL that are most relevant to our article. For an in-depth motivation and extensive hints to relevant literature, we refer the reader to~\cite[Section 1]{RodemannIAL}.

\subsection{Semi-Supervised Learning (SSL)}

The fundamental idea of SSL is to use both labeled data $\mathcal{D}= \left\{ (x_i, y_i) \right\}_{i=1}^n$ and unlabeled data $\mathcal{U}= \left\{ (x_i, \mathcal{Y}) \right\}_{i=n+1}^m$ for training a classification model.\footnote{As such, SSL can be perceived as a special case of superset learning~\cite{hullermeier2014learning,hullermeier2019learning,rodemann2022supersetlearning}.} In this approach, a limited set of labeled data points is combined with an often considerably larger set of unlabeled data points $m >> n$ to train a model $g(x) = \hat{y} \in \mathcal{Y}$, see~\cite{ChapelleSSL}. The incorporation of unlabeled data when training a model can intuitively be useful only if the fundamental SSL assumption is met, i.e., if the unlabeled data contains information about the relationship between the input data and the label of the assigned class~\cite{Zhu}.
In this paper, we will address \textit{classification problems} with self-training methods and incremental addition of pseudo-labeled data as categorized by~\cite{IsaacSL}. A self-training algorithm will, in the first phase, estimate a model from the labeled training data using supervised learning. Based on this model, labels are then predicted for the unlabeled data. These labels are referred to as \textit{pseudo-labels}. In the second phase, the training data is enlarged by adding a selected pseudo-labeled data point according to some criterion. The same point is then removed from the unlabeled dataset, resulting in a new unlabeled dataset. This process is repeated until a stopping criterion is met. This paper will focus on the selection of a potentially favorable pseudo-labeled data point. That is, we take the predicted pseudo-labels as given and only deal with selecting one pseudo-labeled data point from the set of pseudo-labeled data.

\subsection{Bayesian Analysis} In the entire process of PLS we encounter several sources of uncertainty. Traditionally, only the model used for predicting pseudo labels is utilized. Thus, the selection is performed based on the estimated model parameters, without considering the uncertainty associated with those. 
 In the endeavor to establish a selection criterion that is robust against the initial model, i.e., more robust against the choice of model parameters, but at the same time efficiently utilizes the information from labeled data, the choice of a Bayesian approach appears fruitful~\cite{RodemannBPLS,bayesdata24}. Here, it is possible to decompose epistemic uncertainty into statistical uncertainty due to a lack of data and modeling uncertainty due to parametric assumptions. The fundamental idea of Bayesian pseudo-label selection (BPLS) is to choose an instance with a pseudo-label that is probable given the training data but not necessarily probable concerning the estimated model parameters. 
In the spirit of~\cite{reciprocal2024,rodemann2023pseudo,rodemann2022not}, we formalize PLS as a decision problem. The utility of a selected data point $(x_i, \mathcal{Y})$ shall now be the likelihood of being generated jointly with $\mathcal{D}$ by a model with parameters $\theta \in \Theta$ if we include it with the pseudo-label $\hat{y}_i \in \mathcal{Y}$ (obtained through any predictive model) in $\mathcal{D} \cup (x_i, \hat{y}_i)$.
\begin{definition}
\label{definition}
Let $(x_i,\mathcal{Y})$ be any decision (selection) from $\mathcal{U}$. We assign utility to each $(x_i,\mathcal{Y})$ given $\mathcal{D}$ and the pseudo-label $\hat{y} \in \mathcal{Y}$ by the following measurable utility function which is said to be the pseudo-label likelihood.
\begin{equation}
\begin{split}
&u: \mathcal{U} \times \Theta \rightarrow \mathbb{R}\; ,\;((x_i, \mathcal{Y}), \theta) \mapsto 
p( \mathcal{D} \cup (x_i, \hat{y}_i) | \theta) 
\end{split}
\label{Definition 1}
\end{equation} 

\end{definition}
The function $u$ is a natural choice to assign utilities to pseudo-labels given the predicted pseudo-labels. Moreover, under a prior $\pi$, it can be shown that maximizing expected utility, i.e., applying the standard Bayes-criterion $
\Phi(\cdot,\pi): \mathcal{U}  \rightarrow \mathbb{R}; \;            
 a \mapsto \Phi(a, \pi)) = \mathbb{E}_\pi(u(a, \cdot))
$, with respect to $u$ and $\pi$, corresponds to maximizing the pseudo marginal likelihood $p(\mathcal{D} \cup (x_i,\hat{y}_i))$~\cite{RodemannBPLS}.
For any prior, the action which results in the maximal marginal likelihood is Bayes optimal. If we now update the prior with data to a posterior, the standard Bayes criterion corresponds to the pseudo posterior predictive (PPP), i.e., the posterior predictive distribution of the predicted pseudo-labels. 

\subsection{Uncertainty in BPLS}

An important source of uncertainty is the choice of the model, i.e., the distributional assumption under which the likelihood is calculated. So far, this distributional assumption is  the same as the one used for pseudo-label prediction. For addressing uncertainty regarding model choice, a multi-model approach can be applied, which computes a weighted sum of likelihoods of multiple models, see~\cite{RodemannIAL}. 
Additionally, in the same paper, uncertainty regarding the prior selection was briefly discussed. The uncertainty associated with the prior will be thoroughly examined in the remainder of the paper through an implementation via an approximation and experiments.

\section{Related work }

The robustness of pseudo-label selection has been extensively debated in the semi-supervised learning literature. Aminian et al.~\cite{aminian2022information} propose an information theory based PLS approach adaptable to distributional shifts. Vandewalle et al.~\cite{vandewalle2013predictive} aim to enhance model robustness against presumptive biases by facilitating model choice via the deviance information criterion within a semi-supervised learning framework. Rizve et al.~\cite{rizve2020defense} introduce an uncertainty-aware pseudo-label selection technique that competes with popular semi-supervised learning (SSL) methods anchored in consistency regularization. This technique involves selecting pseudo-labels by assessing both their probability scores and predictive uncertainty, the latter gauged by the prediction variance.
Principled approaches, however, that hedge against misspecification of prior assumptions on which the PLS is based upon, are lacking so far. In the context of self-supervised learning, there is the notable exception of Lienen et al.~\cite{lienen2021credal}. Contrary to our approach, they use credal sets for labeling directly rather than for selecting pseudo-labels. Similar to us, they minimize the reliance on distributional assumptions.
Finally, algorithms for computing optimal $\Gamma$-maximin decisions (or optimal decisions with respect to other imprecise decision criteria) have also been discussed, e.g., in~\cite{Utkin:Augustin:2005:power,Jansen2017,jsa2018,festschrift} in the context of decision making under ambiguity or in~\cite{Noubiap2001AnAF,JMLR,pmlr-v216-jansen23a} in the context of robust (Bayesian) statistics. However, these approaches are not (directly) applicable here, as they assume finite state spaces or strictly restricted action spaces of their underlying decision problems, respectively. 

\section{Generalizied Bayesian PLS under Soft Revision}
In real-world applications, the unique choice of an adequate prior as required for Bayesian PLS~\cite{RodemannBPLS} (and any other Bayesian method in machine learning and statistics~\cite{fortuin2022priors,rodemann2022accounting,rodemann2021accounting,rodemann2024explaining,marquardt2023empirical,caprio2023imprecise,imprecise-BO}) is often not straightforward. Instead, several different candidate priors occur equally plausible. In order to make the selection of the pseudo-labeled data more robust against a potentially bad choice of prior in such situations, \textit{credal sets} can be used. A credal set $\Pi$, as first introduced in~\cite{l1974} and heavily been applied in the imprecise probability (IP) community since then, is simply a set of probability measures. Each element $\pi \in \Pi$ is interpreted as an equally plausible candidate prior for the application at hand. Consequently, in the case of modeling with credal sets, any method for pseudo label selection should also include the entire credal set instead of artificially reducing uncertainty to classical probability. 

The transition from a precise prior to a credal set (also referred to as \say{generalized Bayes}~\cite{AugustinIIP}) naturally also requires a decision criterion that can process this information. We choose the Gamma-Maximin criterion, whose fundamental idea is to compute the expected utility under a prior distribution which contrasts the likelihood of the estimated model the most. Then, the instance corresponding to the highest expected utility with respect to its \textit{least favorable prior} should be chosen. Even though Gamma-Maximin can be critized for being rather conservative, one can also argue that this very conservatism is often preferable as it safeguards against taking risky actions on the basis of unjustified assumptions~\cite{AugustinIIP}. In our case, it is better to choose a prior that may not align well with the likelihood rather than selecting one that aligns too closely with the likelihood.

The Gamma-maximin method in turn heavily depends on the choice of the credal set. The larger it is, the higher is the tendency to make a decision with complete disregard of the likelihood. On the other hand, a small credal set might not be expressive enough to sufficiently attenuate the likelihood. Therefore, in the following, we aim to restrict the set of all priors over which we calculate an infimum dynamically, i.e., for each iteration separately~\cite{RodemannIAL}. We adopt a credal set updating rule based on Cattaneo's $\alpha$-cut rule with $\alpha \in (0, 1)$, also known as ``soft revision"~\cite{Cattaneo,augustin2021comment}. This updating strategy aims to update only those priors whose marginal likelihood is greater than or equal to $\alpha$ times the corresponding maximum marginal likelihood. Let, therefore, for the likelihood $\mathcal{L}$ and all $\pi \in \Pi$ be $m(\pi) = \int \mathcal{L}(\theta) f_{\pi}(\theta)\, d\theta$ the marginal likelihood, where $f_{\pi}(\theta)$ is the density of $\pi$ with respect to the Lebesgue measure. Then 
$
\Pi_\alpha = \{\pi \in \Pi \mid m(\pi) \geq \alpha \cdot \sup_{\pi}  m(\pi)\}
$
is the credal set of priors that we update to posteriors. The Gamma-Maximin decision criterion with soft revision is now defined as follows:
\begin{equation}
\begin{split}
&\Phi_{\alpha} : \mathcal{U}  \to \mathbb{R}~~,~~ ((x_i, \mathcal{Y})_i) \mapsto \inf_{\pi \in \Pi_\alpha} \mathbb{E}_{\pi}(u(a, \cdot)).
\end{split}
\label{Gamma}
\end{equation}
\noindent$\Phi_{\alpha}$ addresses uncertainty in a general way without completely ignoring the likelihood. The steeper the likelihood, the smaller the credal set at a constant $\alpha$.

To technically implement\footnote[1]{Implementation of the proposed method and reproducible scripts for the experiments are available at: \url{https://github.com/Stefan-Maximilian-Dietrich/reliable-pls}.} the Gamma-Maximin criterion under soft revision, we first establish a model and a distribution assumption.\footnote{The specific choice of model and distribution assumption here is more for the purpose of demonstration. The following can also be applied to other situations. In case of non-normal priors, alternative approximation strategies need to be considered. } Here, we will consider datasets with bivariate labels and perform logistic regression on them. The labels
are assumed to follow a Bernoulli distribution $Y_i \sim B(p_i)$, where $p_i$ represents the response function that connects the linear predictor to the expected value or the parameters of the distributional assumption: $\mathbb{E}[Y_i] = p_i =  {\exp(\eta_i)}/{(1 + \exp(\eta_i))}$. The credal prior, which the Gamma-Maximin criterion is based on in our case, is a set of normal distributions that vary only in their location parameter, precisely
\begin{equation}
    \Pi =\Bigl\{\pi_{\mu} : \pi_{\mu} = \mathcal{N}(\mu, \Sigma) \wedge \mu \in [\underline{\mu_0},\overline{\mu_0}] \times \ldots \times [\underline{\mu_p}, \overline{\mu_p}]\Bigr\}.
    \label{Credal Set}
\end{equation}
To calculate the marginal likelihood, it is not possible to resort to an analytical method. In this paper, we aim to adapt a numerical method that allows us to approximate the marginal likelihood with the help of the Laplace approximation. 
This is possible because the function for which the integral is to be computed is twice continuously differentiable and most of its mass is around its maximum~\cite{Ruli_2016,bleistein1975asymptotic}.
We define $-h(\theta) := \log(\mathcal{L}(\theta) f_{\pi_\mu}(\theta))$ and $V(\theta)$ as the  Hessian matrix of $h(\theta)$.
Let $\tilde{\theta}$ be the argument that minimizes $h(\theta)$. This minimum is unique and global because $h(\theta)$ is convex. 
Now the Laplace approximation of the marginal likelihood\footnote{$d$ is the number of dimensions. The $\pi$ without a subscript is the number $\pi$.} is: 
\begin{equation}
 m(\pi_\mu) \approx (2\pi)^{\frac{2}{d}} |V(\tilde{\theta})|^{- \frac{1}{2}}\mathcal{L}(\tilde{\theta}) f_{\pi_\mu}(\tilde{\theta}),
\label{Laplace}
\end{equation} \\
see~\cite{Ruli_2016,bleistein1975asymptotic}. To calculate $\tilde{\theta}$ the Broyden–Fletcher–Goldfarb–Shanno (BFGS) algorithm is used~\cite{BROYDEN}. The BFGS method belongs to the quasi-Newton methods and is a numerical technique used to solve nonlinear optimization problems. This method appears particularly suitable for finding optima in our case, as our functions are twice differentiable and it saves computing resources by not computing the Hessian matrix in each step.
To execute the soft revision we now want only the prior $\pi_\mu$ in our credal set that fulfills
\begin{equation} 
\label{Constraint}
m(\pi_\mu) - \alpha \max_{{\pi_\mu} \in \Pi}m({\pi_\mu}) \geq 0
\end{equation}
We now want to construct an approximation of the pseudo posterior predictive (PPP), which depends only on the given data and on the prior $\pi_\mu$. As described in~\cite{RodemannBPLS}, the PPP can be written as:
\begin{equation}
\frac{1}{p_{\pi_\mu}(\mathcal{D})} \int_\Theta  p(\mathcal{D} \cup (x_i, \tilde{y}_i) | \theta) p(\mathcal{D} | \theta)f_{\pi_\mu}(\theta) d\theta
\label{PPP}
\end{equation} 
Here, we can approximate the marginal likelihood $p_{\pi_\mu}(\mathcal{D})$ again using the method described above because the difference is in $\mathcal{O}(n^{-1})$~\cite{RodemannBPLS}. The remaining integral is now similar to the marginal likelihood and can be approximated by Laplace because it is also twice continuously differentiable and most of its mass is around its mode. The approach here is analogous to the one used in approximating the marginal likelihood. Let now
$
 q(\pi_\mu) \approx \int_\Theta  p(\mathcal{D} \cup (x_i, \tilde{y}_i) | \theta) p(\mathcal{D} | \theta)f_{\pi_\mu}(\theta) d\theta.
$
The PPP dependent on the prior $\pi_\mu$ can now be approximated by the approximations of the marginal likelihood and the remaining integral:
\begin{equation}
 p_{\pi_\mu}(\mathcal{D} \cup (x_i, \tilde{y}_i) | \mathcal{D}) \approx \frac{q(\pi_\mu)}{m(\pi_\mu)}
 \label{Approximation}
\end{equation} 
Now we have both a PPP for which it holds 
$
\mathbb{E}_{\pi_\mu}(u(a, \theta)) = p_{\pi_\mu}(\mathcal{D} \cup (x_i, \tilde{y}_i) | \mathcal{D})
$
dependent on a prior or its parameters and a soft revision ($\alpha$-cut) criteria that sorts out all priors that would overly restrict the influence of the likelihood. We now want to treat this as an optimization problem: \\
\begin{equation}
\begin{split}
&\inf_{\pi_\mu \in \Pi} \mathbb{E}_{\pi_\mu}(u(a, \theta))\\  
& \text{s.t. } m(\pi_\mu) - \alpha \max_{\pi_\mu \in \Pi}m(\pi_\mu) \geq 0,
\end{split}
\label{soft_revision}
\end{equation}
which yields the Gamma-Maximin criterion with soft revision, see Equation \ref{Gamma}. \\
The optimization is implemented using the COBYLA (Constrained Optimization by Linear Approximations) algorithm~\cite{Powell}. The basic idea is forming linear approximations to the objective function via interpolation at the vertices of a simplex. This algorithm iteratively proposes new vertices for this simplex, similar to the Nelder-Mead algorithm. The COBYLA algorithm does not rely on gradients and handles inequality constraints well, which is useful for optimization of Equation (\ref{soft_revision}). As termination criterion, we use the fractional tolerance. 

\section{Experiments}
We will now conduct various experiments. For this purpose, we estimate a model in each iteration, which we then evaluate using a test dataset. Since in each round the respective criteria need to be estimated for all pseudo-labeled data, the complexity lies in $\mathcal{O}((m-n)^2)$ with $m-n$ the number of unlabeled data. 
%
We benchmark semi-supervised logistic regression with the Gamma-Maximin method under soft revision ($\alpha$-cut) (\textcolor[HTML]{E3000F}{\textbullet}) against the following methods: 1.) A standard supervised (\textcolor[HTML]{999933}{\textbullet}) baseline where only the labeled data is considered and no self-training takes place. 2.) three standard pseudo-label selection criteria: Probability Score (\textcolor[HTML]{44AA99}{\textbullet}), Predictive Variance (\textcolor[HTML]{000080}{\textbullet}) and Likelihood (max-max) (\textcolor[HTML]{88CCEE}{\textbullet}). 3.) Four (robustified) Bayesian methods that depend on the pseudo posterior predictive: PPP (Bayes-optimal) (\textcolor[HTML]{CC6677}{\textbullet}), PPP multi-label (\textcolor[HTML]{DDCC77}{\textbullet}), PPP multi-model (\textcolor[HTML]{117733}{\textbullet}) and PPP weighted multi-label (\textcolor[HTML]{AA4499}{\textbullet}).\footnote {They incorporate uncertainty regarding model choice (multi-model and weighted multi-model (\textcolor[HTML]{117733}{\textbullet})) and accumulation of errors (multi-label (\textcolor[HTML]{DDCC77}{\textbullet})). For a thorough introduction, please refer to~\cite{RodemannIAL}.}  Experiments are run on simulated binomially distributed data as well as on three data sets (Banknote, MTcars, Mushrooms) for binary classification from the UCI repository~\cite{Dua}. The binomially distributed data was simulated through a linear predictor consisting of normally distributed features. Furthermore, the effect of varying $\alpha$ is tested. Further results can be viewed in the repository.\footnote{\url{https://github.com/Stefan-Maximilian-Dietrich/reliable-pls}.} 
		
		\begin{figure}[!ht]
			\begin{center}
			\includegraphics[width=\textwidth]{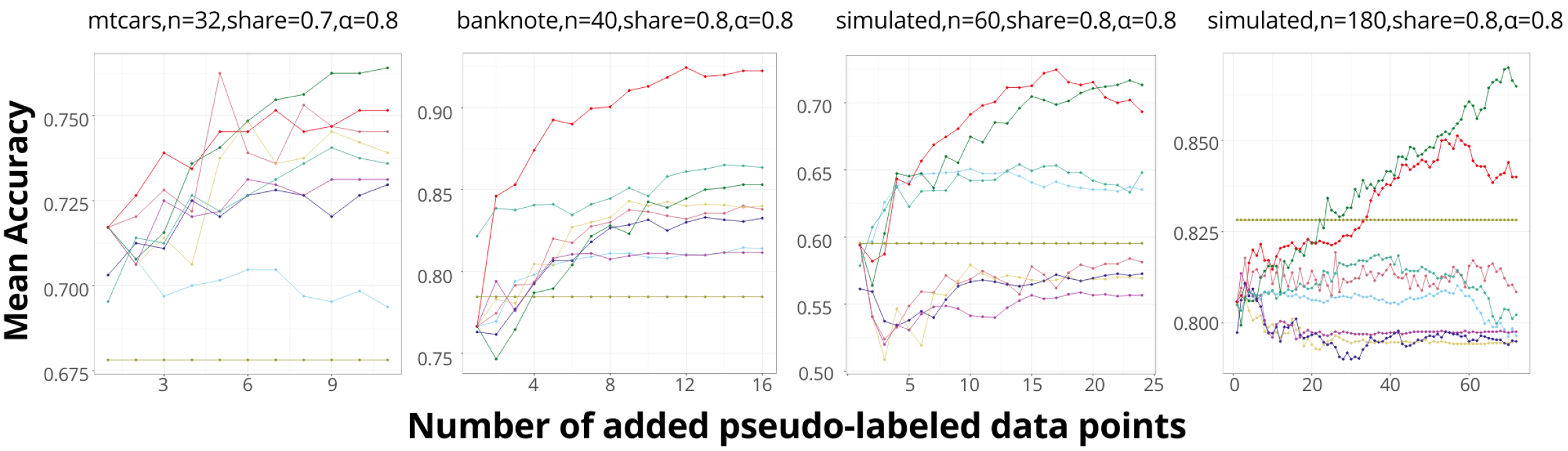}   
			\end{center}
			\caption{Experiments on real-world data (left) and simulated data (right).}
			\label{fig:1}   
		\end{figure}

\section{Discussion}
At first glance, it is apparent that if any of the nine tested methods achieves an improvement over supervised learning, then Gamma-Maximin method with soft revision also achieves an improvement. In the conducted tests, it is evident that the Gamma-Maximin method with soft revision consistently ranks among the best out of all nine tested methods. This is particularly true for real-world data. In the case of simulated data, we see that besides the Gamma-Maximin method with soft revision, only the Multi-Model method consistently performs better than the Supervised Learning baseline. Upon closer comparison, two effects emerge: 1) The lower the total data volume, the better the performance of the Gamma-Maximin method with soft revision. 2) The further away the alternative models are to reality, the better the Gamma-Maximin method with soft revision performs relative to the Multi-Model method (in the example shown here, the alternative models are intentionally chosen to be very favorable to Multi-Model). Additionally, it was observed in the experiments that the Gamma-Maximin method with soft revision copes particularly well with a low proportion of labeled data compared to other methods. In the conducted experiments, no clear trend can be observed regarding which $\alpha$ leads to particularly good performance. 
\printbibliography
\end{document}